\pdfoutput=1
\documentclass{article} %
\usepackage{preprint,times}

\usepackage{hyperref}
\hypersetup{hidelinks}
\usepackage{url}
\usepackage{booktabs}
\usepackage{multirow}
\usepackage{array}
\usepackage{amsmath,amssymb}
\usepackage{graphicx}
\usepackage{xcolor}

\newcommand{\ours}{MeshCarve} %

\newcommand{\cdl}[0]{CD$_{\ell_2}$}
\newcommand{\cdo}[0]{CD$_{\ell_1}$}
\newcommand{\hd}[0]{HD}
\newcommand{\nc}[0]{NC}

\title{\ours: Artisan Mesh Generation with Flow Matching in Compact Latent Spaces}

\author{Xiyu Wang$^{1*}$, \enspace Ruocheng Wu$^{2*}$, \enspace Yufei Wang$^{2\dagger\ddagger}$, \enspace Zhihao Li$^{1,2}$, \enspace Lanqing Guo$^{3}$, \enspace Bihan Wen$^{1\dagger}$ \\[4pt]
{\normalfont $^1$Nanyang Technological University \quad $^2$SparAI Inc. \quad $^3$University of Texas at Austin} \\
{\normalfont\small $^*$Equal contribution \quad $^\dagger$Corresponding authors \quad $^\ddagger$Project lead}}

\preprintfinalcopy
\renewcommand{\headrulewidth}{0pt}
\begin{document}

\maketitle

\begin{center}
\includegraphics[width=\linewidth]{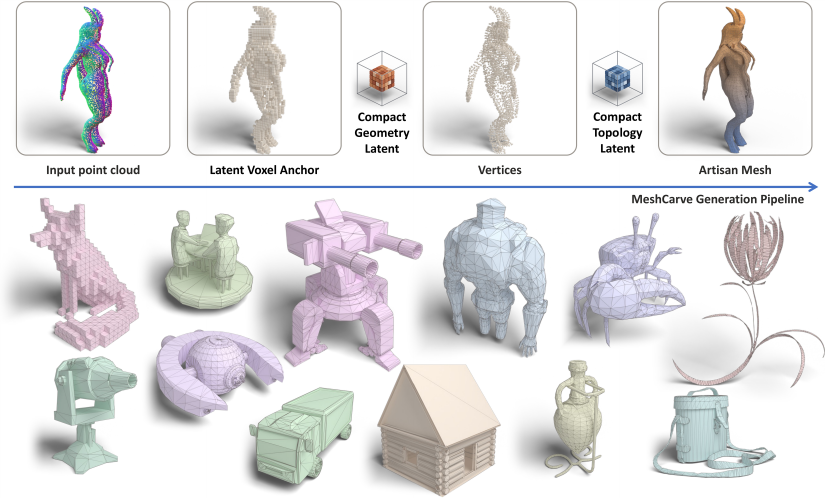}
\end{center}
\refstepcounter{figure}\label{fig:teaser}
{\small\textbf{Figure \thefigure:} \ours{} turns a dense point cloud into an artisan mesh with two flows on a generated anchor. The latent voxel anchor is generated with a variational autoencoder. The vertex flow then places the vertices, and the edge flow decides the connectivity between vertices. Both flows run in a spatially compressed $64^3$ latent, drawn as the two compact cubes. Bottom: meshes generated by \ours{}.}
\vspace{8pt}

\begin{abstract}
Prior artisan mesh generation works largely predict face tokens autoregressively, which makes inference slow. Recent methods instead flow match continuous latents built by Variational AutoEncoders (VAEs), but reconstruction quality drops significantly when geometry and topology are jointly encoded, and further when the latent space is compressed. We present \textbf{MeshCarve}, a flow matching method that generates entirely in compact latent spaces, generating vertex positions and edge connections separately and sidestepping the difficulty of a joint compact latent. To shorten the token sequence, we propose a hierarchical sparse transformer backbone, instantiated as VertexVAE and EdgeVAE. Instead of encoding fields over the surface voxels, both VAEs anchor on discrete vertices in their latent spaces, which drastically reduces the token sequence length, and our spatial-aware compression shortens it further without costing reconstruction. VertexVAE directly encodes vertex occupancy. For connectivity, we propose vertex-link encoding, which turns arbitrary connectivity between vertices into fixed-length continuous per-vertex embeddings and recovers complex artistic topology faithfully. MeshCarve combines these VAEs with an anchor generator and flow matches on the shortened token sequences. It shows advantages over state-of-the-art autoregressive and flow matching methods on Objaverse and generalizes to Toys4K. To the best of our knowledge, it is among the first artisan mesh generation methods whose every generative stage runs in a spatially compressed latent, with a token sequence only a fraction of the most compressed previous autoregressive and flow matching works.

\end{abstract}

\section{Introduction}
\label{sec:intro}
3D mesh models are widely used as digital representations of 3D shapes. They are used in applications such as video games, visual effects, animation, and virtual reality. In practice, artisan meshes handcrafted by human artists with well-structured topology are particularly valuable: their compact geometry supports efficient rendering, while their carefully designed topology facilitates downstream tasks such as editing, deformation, and texture mapping. Hence, generating meshes with artistic geometry and topology is a central problem for 3D content creation~\citep{meshgpt,meshanything,bpt,deepmesh,lato1,meshflow}.

Many works model mesh generation as autoregressive generation of face tokens with transformers. Faces are tokenized and predicted one by one, starting from vertex-coordinate tokens~\citep{polygen,meshgpt} and progressing through increasingly compact tokenizations that share vertices across adjacent faces~\citep{meshanythingv2}, walk edges~\citep{edgerunner}, or index blocks and patches~\citep{bpt,deepmesh}. These efforts have greatly reduced the token budget for a given mesh, and the most recent methods reach one token per face~\citep{face}. Diffusion and flow matching methods~\citep{polydiff,meshcraft,spacemesh} show greater potential, as they sidestep serialized generation, which is slow by nature. Among them, two recent works anchor the latent to the sparse structure of the mesh, one token per surface voxel~\citep{lato1} or per vertex~\citep{meshflow}, and reach quality competitive with autoregressive generation. Both carry geometry and topology in one latent, and neither compresses it spatially. The surface-anchored latent keeps every surface voxel, and the vertex-anchored one reports a large reconstruction loss when its tokens are spatially down-sampled. The flow transformer therefore attends over a sequence that grows with the mesh, and its cost grows quadratically with it. \emph{We ask whether flow matching can instead run entirely in a spatially compressed latent.} The token budget would then grow only sublinearly as the meshes size up.

The surface-anchored and the vertex-anchored latent each have further limitations that motivate our design.
First, both jointly encode geometry and topology in a shared latent representation. Echoing the known limitations of shared latent spaces in multimodal VAEs~\citep{daunhawer2022limitations,javaloy2022modality}, our preliminary experiments show that such joint encoding substantially degrades the reconstruction quality of prior work~\citep{lato1}.
Second, neither representation in its existing form is well suited for further compression. A surface-anchored latent assigns tokens to every surface voxel, even though surface voxels are several times more numerous than vertex-containing voxels. In existing vertex-anchored designs, topology is represented using a fixed-length ordered list of neighbors for each vertex, making the representation sensitive to neighbor ordering and imposing an upper bound on vertex degree. Consequently, the former contains substantial spatial redundancy, whereas the latter cannot merge vertex tokens without risking the loss of essential connectivity information. Importantly, this limitation arises from how topology is represented at the vertices, rather than from vertex anchoring itself.

MeshCarve addresses each of these limitations. Given an input point cloud, it first generates the coarse vertex occupancy with a VAE in a single forward pass and then generates vertex positions and edge connections sequentially through flow matching, each in its own spatially compressed latent space. These latent spaces are learned by VertexVAE and EdgeVAE, respectively, which use the same hierarchical sparse transformer VAE architecture built around a redesigned vertex-anchored representation. The architecture compresses the latents by merging the eight children of each voxel through a learned projection, allowing the token count to adapt to the local vertex density. For connectivity, EdgeVAE uses vertex-link encoding to embed the links from each vertex to its neighbors into a fixed-width, order-invariant representation, from which its decoder recovers edges through classification. Recent and concurrent works~\citep{nexus,lato2} also separate geometry from topology. MeshCarve further compresses both latents, preserving the reconstruction benefit of separation while shortening both generative sequences. Evaluated against artist references, MeshCarve shows advantages over prior state-of-the-art methods on Objaverse and generalizes to Toys4K.

In sum, our contributions are threefold:
\begin{itemize}
\item We propose \ours{}, which samples the coarse vertex occupancy with a VAE in one forward pass and then flow matches vertex positioning and edge connection in turn, each in its own spatially compressed latent.
\item We propose a hierarchical sparse transformer VAE backbone with \textbf{spatial-aware compression}, instantiated as the VertexVAE and the EdgeVAE, and the \textbf{vertex-link encoding}, which turns arbitrary connectivity into a fixed-length, order-free per-vertex embedding.
\item The latent holds 0.13 tokens per face on average, the shortest among the compared methods. That budget affords a larger flow transformer, and \ours{} shows advantages over previous state-of-the-art methods on Objaverse and generalizes to Toys4K.
\end{itemize}

\begin{table}[t]
\centering
\caption{Tokens per triangle face, lower is a shorter sequence for the generative transformer. Target: G = geometry, T = topology, joint = one latent for both. Two-stage methods list both stages, and the two stages of \ours{} share the same anchor and hence one value.}
\label{tab:tokens}
\footnotesize
\setlength{\tabcolsep}{4pt}
\resizebox{\textwidth}{!}{%
\begin{tabular}{lc@{\hspace{12pt}}lllc}
\toprule
\multicolumn{2}{c}{Autoregressive} & \multicolumn{4}{c}{Diffusion / Flow Matching} \\
\cmidrule(lr){1-2}\cmidrule(lr){3-6}
Method & Tokens / face & Method & Token & Target & Tokens / face \\
\midrule
MeshAnything~\citep{meshanything} & 9 & LATO~\citep{lato1} & coarse surface voxel& joint & 6.5 \\
MeshGPT~\citep{meshgpt} & 6 & LATO.2~\citep{lato2} & coarse surface voxel / vertex& G+T & 0.88 / 0.32 \\
EdgeRunner~\citep{edgerunner} & 4.2 & MeshFlow~\citep{meshflow} & vertex & joint & 0.52 \\
MeshAnything V2~\citep{meshanythingv2} & 4.1 & Nexus~\citep{nexus} & vertex & G+T& 0.79 / 0.52\\
DeepMesh~\citep{deepmesh} & 2.5 & MeshCraft~\citep{meshcraft} & face & joint & 1.0 \\
BPT~\citep{bpt} & 2.3 & \ours{} (Ours)& coarse vertex voxel& G+T & \textbf{0.13} \\
TreeMeshGPT~\citep{treemeshgpt} & 2.0 & & & & \\
FACE~\citep{face} & 1.0 & & & & \\
\bottomrule
\end{tabular}}
\end{table}

\section{Related Work}
\label{sec:related}
\textbf{Autoregressive artisan mesh generation.}
PolyGen~\citep{polygen} and MeshGPT~\citep{meshgpt} established the face-token paradigm: faces are tokenized and a transformer predicts them autoregressively. Subsequently, MeshAnything~\citep{meshanything} scales up based on them and tokenizes each face with 9 tokens, i.e., the 3D coordinates of its three vertices. Due to this inefficient tokenization, it is limited to generating meshes with a few hundred faces. Subsequent works~\citep{meshanythingv2,edgerunner,bpt,deepmesh,treemeshgpt} improve the token efficiency progressively. Recent efforts~\citep{face} reach one token per face with an MLP tokenizer. Nevertheless, a few challenges remain. First, artisan meshes frequently contain thousands or tens of thousands of faces. At one token per face, the token sequence is still unaffordably long and the generation can take very long. Second, the serialized generation is prone to accumulated errors and often fails to emit a mesh with complete geometry because the stopping token is emitted too early. MeshMosaic~\citep{meshmosaic} thus proposes to generate block by block to decompose the generation process, but these generic limitations still apply.

\textbf{Latent generative models for artisan meshes.}
Flow matching and diffusion models sidestep the serialization of autoregressive generation, since vertices, edges and faces are produced concurrently. The pioneering works PolyDiff and SpaceMesh~\citep{polydiff,spacemesh} apply diffusion to triangle soups and to raw vertex coordinates, respectively, and MeshCraft~\citep{meshcraft} flow matches a latent of independent faces. Later works mix the paradigms. FastMesh~\citep{fastmesh} generates the vertex positions autoregressively and obtains the connections from a predictor in one shot. TriFlow~\citep{triflow} generates a topology field over a given surface and takes its vertices from that surface. Nexus~\citep{nexus} generates the vertices by diffusion in raw voxel space, as subdivision and pruning, and then diffuses in latent space to generate the connections. In 3D shape generation, sparse voxel VAEs compress high-resolution surfaces into compact latents for diffusion and flow matching~\citep{trellis,sparc3d,trellis2}. Notably, LATO~\citep{lato1} and MeshFlow~\citep{meshflow} flow match in a continuous latent built by a Variational AutoEncoder (VAE) that encodes vertex positions and connections jointly. LATO anchors the latent to the surface voxels and MeshFlow to the vertices, and both reach quality competitive with autoregressive generation. Both also keep one token per element, so the latent grows with the mesh and flow matching in it is costly. Their joint encoding of vertex positions and connections further degrades reconstruction. In \ours{}, we anchor flow matching to vertices and construct compact latent spaces that reduce the token sequence length, allowing a larger flow matching backbone to run comfortably.

\section{Method}
\label{sec:method}
\begin{figure}[t]
\centering
\includegraphics[width=\linewidth]{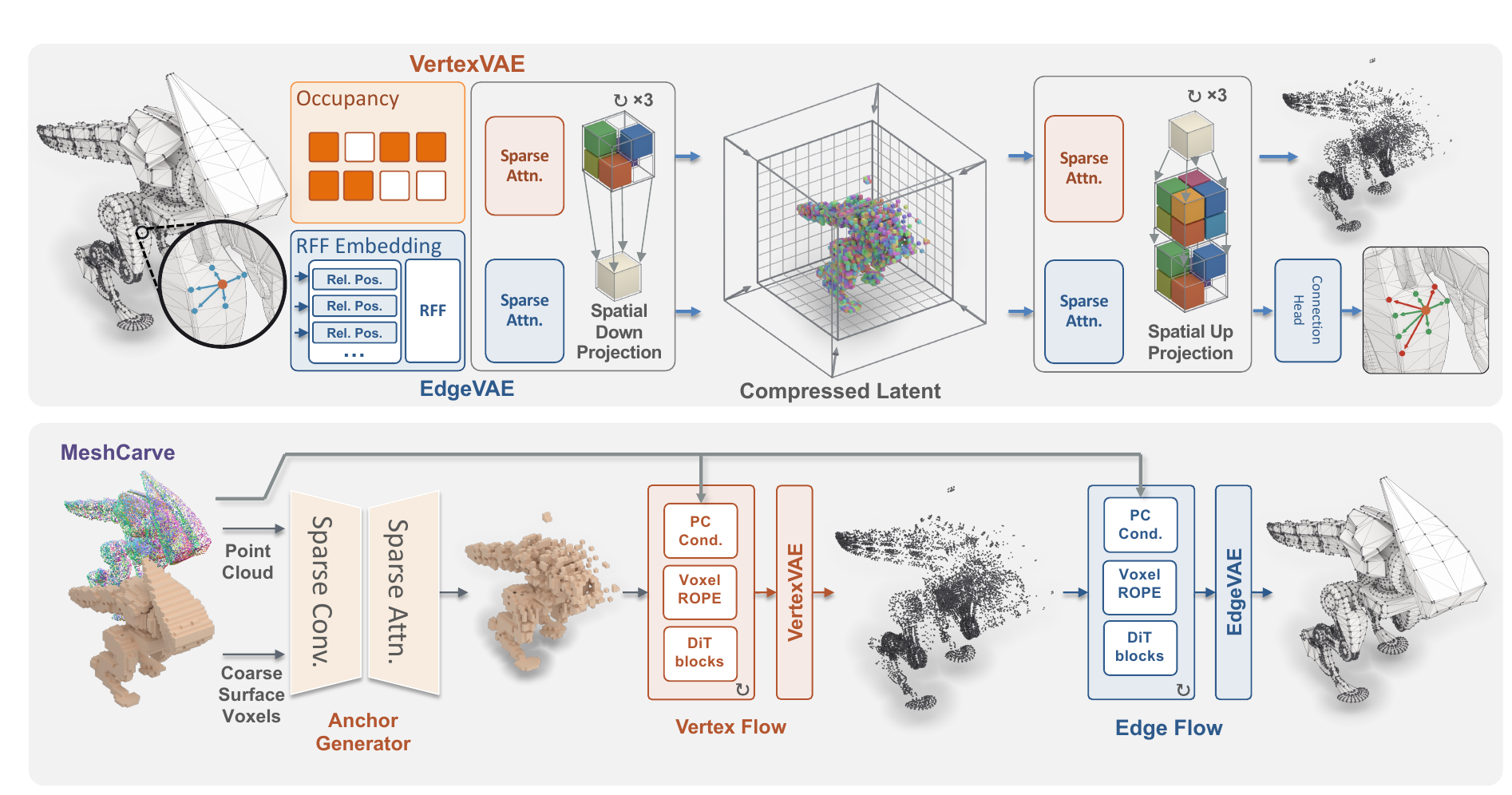}
\caption{Overview of \ours{}. Top: VertexVAE and EdgeVAE share one sparse transformer backbone, reaching the compressed $64^3$ latent through the spatial down projection of Eq.~\ref{eq:s2c} and decoding the vertex voxels and their connections through the mirrored up projection. Bottom: end-to-end generation from a dense point cloud through the anchor generator and two flows.}
\label{fig:method}
\end{figure}

\ours{} generates an artisan mesh in two stages (Figure~\ref{fig:method}). We propose an anchor generator that settles the latent voxel positions for flow matching. The vertex flow then generates a compact latent on the anchor, which decodes into the fine vertex occupancy. The edge flow generates a second compact latent on the same anchor, and this latent decodes into the connectivity. Both latents are encoded using the same VAE architecture, instantiated as the VertexVAE and the EdgeVAE. We describe the backbone and the two VAEs first and the pipeline second.

Throughout, we write a mesh as a pair $(\mathbf{V},\mathbf{A})$, with $\mathbf{V}\in\mathbb{R}^{N\times 3}$ the $N$ vertex positions $\mathbf{v}_i$ in the unit cube and $\mathbf{A}\in\{0,1\}^{N\times N}$ the symmetric adjacency, $A_{ij}=1$ when vertices $i$ and $j$ share an edge. Faces are triangular, not stored, and recovered as the 3-cycles of $\mathbf{A}$. Discretizing the unit cube into an $R^3$ grid records which voxels contain vertices as a binary occupancy $\mathbf{S}_R$, and an occupied voxel is a vertex voxel. Both stages of \ours{} follow the latent-diffusion recipe~\citep{ldm}, a VAE~\citep{vae} mapping the mesh to a compact latent whose posterior is kept near a standard Gaussian and a flow model trained in that space. The reconstruction term of each VAE is given below and the flow objective in Appendix~\ref{app:flow}.

\subsection{Overall VAE Design}
\label{sec:vae}

Flow matching needs the mesh as a continuous latent, and a mesh carries two signals of different kinds: discrete vertex positions and combinatorial connectivity. Prior works such as LATO~\citep{lato1} and MeshFlow~\citep{meshflow} choose to encode both into one latent, costing reconstruction quality. Our preliminary studies also show similar failure modes. Notably, this is a known issue in the field of multi-modal VAEs~\citep{daunhawer2022limitations} and modality-specific latents recover the quality~\citep{palumbo2023mmvaeplus}. We therefore encode vertex position and edge connection with separate VAEs that share one backbone.

To design the VAE, we point out that vertices are the fewest of a mesh's primitives (around 12-16\% of surface voxels) and naturally fit into quantized voxel grids, so we anchor our VAEs to vertex voxels. We also observe that the key to running flow matching at a larger scale is to compress the latent space and shorten the token sequence, but current generation works~\citep{meshflow} encounter a noticeable reconstruction barrier when compressing the latent space. To this end, we propose to compress spatially through a learned projection rather than by fixed-interval pooling~\citep{meshflow}. Notably, we found that convolutional networks operate inefficiently on vertices, which are highly sparse. Thus, our VAE is fully based on sparse transformers that assign 3D positional embeddings to sparse voxels.

\paragraph{Spatial-Aware Compression}
In recent years, with the rapid improvement of hardware, e.g., GPUs, applied artisan meshes are also growing larger. With vertex-anchored tokens, we can shorten the token sequence considerably, but it remains difficult to afford a larger flow matching network because the vertex-anchored token sequence still grows linearly as the demanded mesh sizes up. Worse, the cost of running mainstream DiT networks grows quadratically with the token sequence length. MeshFlow~\citep{meshflow} proposed TokenMerge, which merges tokens that are consecutive in the vertex index and projects them down. Nevertheless, this compression trades reconstruction significantly. To this end, we point out that compression should attend to the spatial relationship between voxels and propose spatial-aware compression. Specifically, inspired by \citet{dcae,trellis2}, eight children of a parent voxel are stacked channel-wise and then a linear mapping is learned to reach the dimension of a single voxel, reaching a coarser level.
The spatial-aware compression and its mirrored decompression are defined as
\begin{equation}
\mathbf{h}_c=\sum_{k=1}^{8}\mathbf{W}^{\downarrow}_{k}\,\mathbf{h}_{c,k},\qquad
\mathbf{f}_{c,k}=\mathbf{W}^{\uparrow}_{k}\,\mathbf{f}_c,\quad k=1,\dots,8,
\label{eq:s2c}
\end{equation}
where $c$ is a parent voxel, $c,k$ its child in octant $k$, $\mathbf{h}$ and $\mathbf{f}$ are encoder and decoder features of width $D$, and $\mathbf{W}^{\downarrow}_{k},\mathbf{W}^{\uparrow}_{k}\in\mathbb{R}^{D\times D}$ are learned per octant. Empty children drop out of the sum on the way down, and only occupied children are instantiated on the way up. The compression is therefore spatial-aware, and the token count follows the local vertex density rather than the grid resolution. In both VertexVAE and EdgeVAE the spatial-aware compression is applied after transformer blocks. It reduces the input resolution of $512^3$ to $256^3$, then $128^3$, and finally $64^3$, which is the resolution of our sparse compact latent. Following standard VAE practice, a linear layer maps each 768-dimensional latent token to the mean and log-variance of a 16-channel Gaussian for sampling.

\subsection{VertexVAE: Compressing Vertex Geometry}
\label{sec:VertexVAE}
The geometry side only has to decide which voxels hold vertices, so its decoder classifies octants rather than regressing coordinates. VertexVAE encodes vertex occupancy at $512^3$ into the $64^3$ latent, and its decoder recovers the vertex voxels by subdivision and pruning, following~\citet{trellis2,lato1}. At each of the three levels, every latent voxel is expanded into its eight child features by the upsampling of Eq.~\ref{eq:s2c}. A prune head then scores each child with the probability that it contains a vertex, $p_{c,k}=\mathrm{sigmoid}(\mathrm{MLP}(\mathbf{f}_{c,k}))$, where $\mathrm{MLP}$ is a small multilayer perceptron. Children with $p<0.5$ are discarded and the survivors subdivided again until the $512^3$ vertex voxels are reached, whose centers are the vertex positions. The prune heads are trained with the asymmetric loss (ASL)~\citep{asl} following~\citet{lato1}. The network is trained with teacher-forcing. The inference is cascaded subdivision and pruning.

\subsection{EdgeVAE: Compressing Vertex Connectivity}
\label{sec:EdgeVAE}
Topology is the set of connections between vertices. It is conventionally a byproduct of generating faces~\citep{polygen,meshgpt,meshanything,bpt}. Recent works~\citep{spacemesh,lato2,nexus} instead generate the adjacency matrix for a given set of vertices directly. Connectivity carries long-range information and is harder to reconstruct faithfully, let alone from a spatially compact latent. EdgeVAE therefore keeps the backbone above (Eq.~\ref{eq:s2c}).

\paragraph{Vertex-Link Encoding.}
To encode connectivity with a single fixed-width token per vertex, the variable-size set of its neighbors must be mapped to a fixed-dimensional representation. A straightforward fixed-width construction assigns neighbor coordinates to ordered slots and pads the list to a prescribed maximum degree~\citep{meshflow}. However, this construction is sensitive to permutations of the neighbor set, imposes a hard cap on vertex degree, and reserves dimensions for uninformative padding. We therefore propose vertex-link encoding, which carries the connections of a vertex as the relative positions of the vertices it is connected to, encoded with random Fourier features (RFF)~\citep{rff} and summed over the neighbors. The encoding is straightforward, with no parameters to learn and no hand-crafted rule beyond one fixed random matrix. It has a fixed length without padding. It is order-free, so two exports of the same mesh give the same token. It accepts any valence, so no capacity is spent on empty slots and no cap is placed on the degree. \citet{tancik} compare frequency mappings for low-dimensional coordinates and find isotropic Gaussian frequencies preferable to on-axis positional encodings, which are biased toward axis-aligned structure. The offsets from a vertex to its linked vertices point in arbitrary directions, so we draw $\mathbf{B}\in\mathbb{R}^{m/2\times 3}$ once from $\mathcal{N}(0,\sigma^{2}\mathbf{I})$ and keep it fixed. Each relative position $\mathbf{d}$ is encoded as $\gamma(\mathbf{d})=[\cos(\mathbf{B}\mathbf{d});\,\sin(\mathbf{B}\mathbf{d})]\in\mathbb{R}^{m}$, and the input token of vertex $i$ sums the encodings of all its connections,
\begin{equation}
\mathbf{r}_i=\sum_{j:\,A_{ij}=1}\gamma(\mathbf{v}_j-\mathbf{v}_i),\qquad
\langle\gamma(\mathbf{a}),\gamma(\mathbf{b})\rangle\approx\tfrac{m}{2}\,e^{-\sigma^{2}\|\mathbf{a}-\mathbf{b}\|^{2}/2},
\label{eq:rff}
\end{equation}
with $m=128$ and $\sigma=3$ in voxel units. The sum has a fixed width whatever the number of connections, and since the inner product of two encodings approximates a Gaussian kernel of their difference, distinct relative positions remain separable inside the sum. During decoding, the vertex voxels are given, so the decoder expands the $64^3$ latent along them with the upsampling of Eq.~\ref{eq:s2c} without pruning and produces a feature for every vertex. Following LATO~\citep{lato1}, a connection head classifies each vertex pair and is trained with binary cross-entropy.

\subsection{Generating Artisan Mesh}
\label{sec:gen}
\subsubsection{Generating the Latent Voxel Anchor}
To begin flow matching, position of the latent voxel should be given. This is an essential step because otherwise the flow matching needs to operate on a dense tensor with hundreds of thousands of tokens. Our preliminary study generated it with a flow matching model following the setup of TRELLIS~\citep{trellis}. However, it costs significant compute and the result remains unsatisfactory (Table~\ref{tab:ablation}). Our insight is that the coarse geometry of an artisan mesh is mostly determined by the surface itself. Vertices gather where the curvature is. At the coarse resolution, such as $64^3$, the different tessellations an artist might choose occupy the same voxels, leaving less ambiguity and making flow matching or diffusion redundant. We therefore generate the anchor in one shot with an anchor generator. Notably, this is faster than flow matching and better for this task. The generator is a sparse VAE composed of convolutional and attention blocks. Its input is the dense point cloud and the coarse surface occupancy. We use a PointNet to pool the position and normals of point cloud samples into the voxels, giving coarse voxels richer details. The vertex budget is also given as a condition since it is up to user preference. The latent space is a dense tensor with a spatial dimension of $16^3$. The decoder subdivides and prunes redundant voxels to generate the placement of the coarse vertex positions, i.e., the anchor that flow matching networks run on. We use the asymmetric loss (ASL)~\citep{asl} since vertices are highly sparse. Additional backbone details can be found in the supplementary.

\subsubsection{Vertex and Edge Latent Space Flow Matching}
Both latents are generated with rectified flow~\citep{rectifiedflow,flowmatching} by the sparse transformer of~\citet{trellis,trellis2}, with one token per latent voxel and the dense point cloud as the condition through a frozen shape encoder~\citep{michelangelo}. The vertex flow runs on the generated anchor and is additionally conditioned on the vertex budget $N$. Its sample decodes into the vertex voxels at $512^3$, whose centers are the generated vertices. The edge flow runs on the same anchor and is additionally conditioned on the vertex latent of the same voxel. Its sample decodes over the generated vertex voxels into the adjacency and the faces through the connection head. Both flows train on the ground-truth anchor, vertex voxels and vertex latents, and run on generated ones at inference. The objective, the sampler and the conditioning are detailed in Appendix~\ref{app:flow}.

\section{Experiments}
\label{sec:exp}
\label{sec:setup}
\paragraph{Implementation}
VertexVAE and EdgeVAE share the sparse transformer backbone, with a hidden width of 768 and eight blocks at each resolution level, at 429M and 483M parameters. The anchor generator has 613M parameters. The vertex and edge flows follow the implementation of TRELLIS~\citep{trellis,trellis2}, with 30 blocks at width 1,536 and 32 blocks at width 2,048, 1.27B and 2.37B parameters. \ours{} is trained on 558K meshes from Objaverse~\citep{objaverse}, with the maximum vertex count capped at 20,000. Generation is evaluated on 185 held-out Objaverse meshes and on 500 meshes drawn at random from Toys4K~\citep{toys4k}, and the VAE ablations on 5,000 Objaverse validation meshes. Training schedules and inference settings are given in the supplementary.

\paragraph{Baselines and metrics}
We compare against MeshAnything, MeshAnything V2, TreeMeshGPT, BPT, DeepMesh, FastMesh, MeshFlow and LATO.2~\citep{meshanything,meshanythingv2,treemeshgpt,bpt,deepmesh,fastmesh,meshflow,lato2}, each run from its released weights under its stock inference setting. Autoregressive methods that publish a face cap are scored on the evaluation meshes inside it. The topology flow of LATO.2 is scored on the meshes within its vertex bound. In each case \ours{} is scored on the same meshes, and on the full sets otherwise. \ours{} and LATO.2 take the reference vertex count as their user-set budget. The supplementary gives the settings of every baseline, the count conditioning, and the comparison outside the caps. LATO~\citep{lato1} released only its VAE. All methods are scored in the same way, following~\citet{meshflow,nexus}. Prediction and reference are normalized to the $[-0.5,0.5]$ box, and 50K points are sampled from each surface. We report Chamfer distance (\cdl{}, \cdo{}), Hausdorff distance (\hd{}) and normal consistency (\nc{}).

\subsection{Main Results}
\label{sec:main}
\label{sec:variance}

\begin{table}[t]
\centering
\caption{Shape-conditioned artisan mesh generation on Objaverse and Toys4K. The top block gives the edge stage the reference vertices. Best in bold within each block.}
\label{tab:main}
\small
\setlength{\tabcolsep}{3.5pt}
\begin{tabular}{lcccc@{\hspace{8pt}}cccc}
\toprule
& \multicolumn{4}{c}{Objaverse} & \multicolumn{4}{c}{Toys4K} \\
\cmidrule(lr){2-5}\cmidrule(lr){6-9}
Method & \cdl{}$\downarrow$ & \cdo{}$\downarrow$ & \hd{}$\downarrow$ & \nc{}$\uparrow$ & \cdl{}$\downarrow$ & \cdo{}$\downarrow$ & \hd{}$\downarrow$ & \nc{}$\uparrow$ \\
\midrule
\multicolumn{9}{l}{\emph{w/ GT vertices}} \\
LATO.2 T-Flow~\citep{lato2} & .00595 & .00824 & .0709 & .8775 & .00483 & .00689 & .0490 & .9245 \\
\ours{} edge flow (ours) & \textbf{.00462} & \textbf{.00644} & \textbf{.0495} & \textbf{.9218} & \textbf{.00374} & \textbf{.00538} & \textbf{.0461} & \textbf{.9488} \\
\midrule
\multicolumn{9}{l}{\emph{End-to-end generation}} \\
MeshAnything~\citep{meshanything} & .02838 & .03655 & .1661 & .7537 & .03525 & .04798 & .1765 & .7754 \\
MeshAnythingV2~\citep{meshanythingv2} & .01907 & .02611 & .1513 & .7867 & .03549 & .04858 & .1683 & .7667 \\
TreeMeshGPT~\citep{treemeshgpt} & .02611 & .03554 & .1659 & .7584 & .01459 & .02061 & .0841 & .8607 \\
BPT~\citep{bpt} & .01561 & .02155 & .0991 & .8406 & .00866 & .01256 & .0617 & .8962 \\
DeepMesh~\citep{deepmesh} & .01991 & .02738 & .1382 & .7833 & .01055 & .01515 & .0705 & .8871 \\
FastMesh~\citep{fastmesh} & .00998 & .01415 & .0922 & .8010 & .00780 & .01128 & .0618 & .8702 \\
\midrule
MeshFlow~\citep{meshflow} & .00854 & .01212 & \textbf{.0547} & .8119 & .00676 & .00978 & \textbf{.0355} & .8793 \\
LATO.2~\citep{lato2} & .00693 & .00993 & .0754 & .8238 & .00562 & .00820 & .0476 & .8855 \\
\ours{} (ours) & \textbf{.00649} & \textbf{.00933} & .0789 & \textbf{.8529} & \textbf{.00560} & \textbf{.00814} & .0666 & \textbf{.9069} \\
\bottomrule
\end{tabular}
\end{table}

As shown in Table~\ref{tab:main}, \ours{} achieves the highest normal consistency on both benchmarks and the lowest Chamfer distance on Objaverse among all compared methods. On Toys4K it is level with LATO.2 in Chamfer distance. As a maximum-based metric, Hausdorff distance can be dominated by a single hole or uncovered region. MeshFlow~\citep{meshflow} achieves the lowest value with coarse, low-polygon outputs that capture the global shape without large local gaps. This robust surface coverage is favored by Hausdorff distance, which does not directly assess tessellation density or edge layout. Additional analysis of vertex count adherence is included in the supplementary, where we find that MeshCarve follows the given vertex count better than MeshFlow and LATO.2. The autoregressive baselines are scored inside their published face caps, the range they were built for, and the gaps widen on the full sets (Appendix~\ref{app:arcap}).

\begin{figure}[t]
\centering
\includegraphics[width=\linewidth]{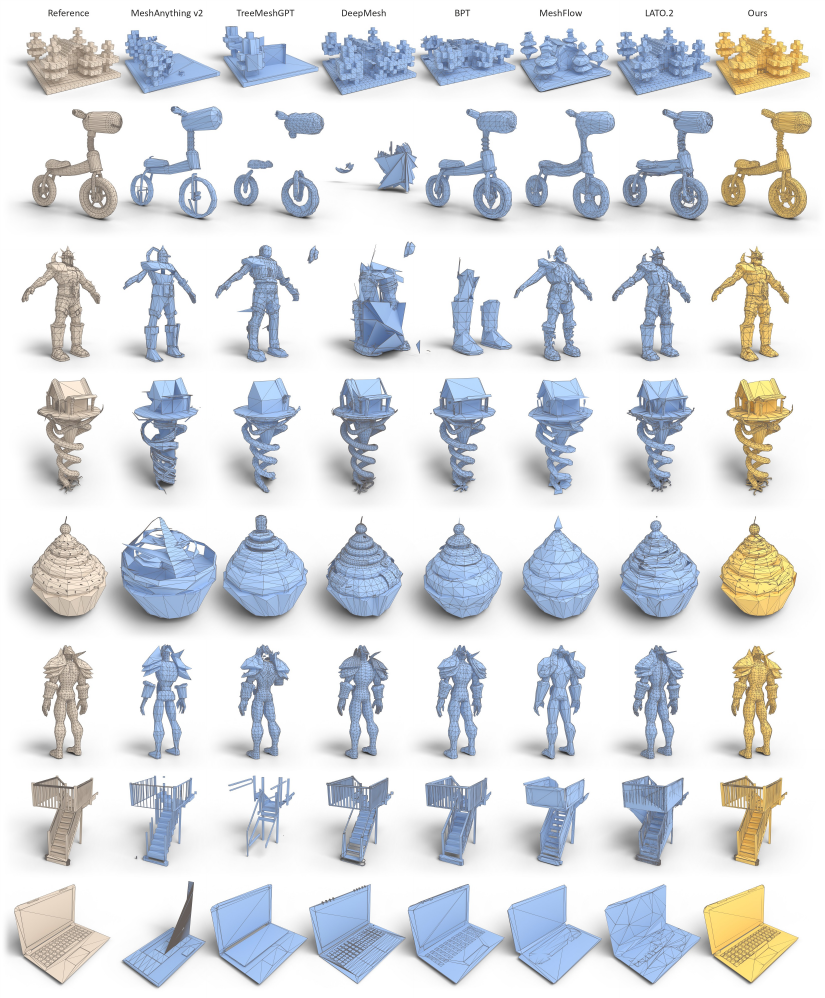}
\caption{Qualitative comparison on Objaverse and Toys4K. References are in clay, baselines in blue and \ours{} in amber; all outputs are shown as generated with wireframes. Zoom in for better view.}
\label{fig:qual}
\end{figure}

We also compare the edge flows alone. Both are given the same reference vertices, on the meshes that the topology flow of LATO.2 admits. Our edge flow leads on every metric (Table~\ref{tab:main}, top block) by significant margins. On Objaverse the lead is 22\% to 30\% of the mean Chamfer and Hausdorff distance. It also recovers more of the artist's wiring, as measured by the edge F1 score (Table~\ref{tab:vae_recon}). We believe the lead comes mostly from the much higher compression rate of our EdgeVAE, which allows flow matching to run at an affordable computational cost. Our VertexVAE and the vertex flow also show stronger performance than the V-VAE of LATO.2 (Table~\ref{tab:vae_recon}). In sum, we demonstrate that both vertex position and edge connectivity can be flow matched in much more compact latent spaces, allowing easy scale-up of the flow matching model. Notably, the V-VAE of LATO.2 does not generalize to our test dataset despite our best effort to reproduce it. Surface metrics do not see the connectivity, so Appendix~\ref{app:struct} reports statistics of the generated graphs, where our edge flow given the reference vertices produces far fewer non-manifold edges than the topology flow of LATO.2.

\begin{table}[t]
\makeatletter
\newsavebox{\tabAbox}\newsavebox{\tabBbox}\newlength{\tabH}
\setbox\tabAbox\vbox{\hsize=0.42\textwidth\@parboxrestore
\caption{\textbf{VAE reconstruction and flow generation.} (a) F1 of the reconstructed vertex voxels and edges. (b) The vertex flow scored against, and the edge flow given, the reference vertices. Best in bold.}
\label{tab:vae_recon}
\vfill
\centering
\scriptsize
\setlength{\tabcolsep}{2.5pt}
\renewcommand{\arraystretch}{1.0}
\begin{tabular}[b]{lcccc}
\toprule
\multicolumn{5}{l}{\emph{(a) VAE reconstruction F1$\uparrow$}} \\
& \multicolumn{2}{c}{Vertex VAE} & \multicolumn{2}{c}{Edge VAE} \\
\cmidrule(lr){2-3}\cmidrule(lr){4-5}
Method & Toys4K & Objaverse & Toys4K & Objaverse \\
\midrule
LATO & .762 & .656 & .866 & .667 \\
MeshFlow & .017 & .020 & .973 & .920 \\
LATO.2 & .375 & .343 & \textbf{.999} & \textbf{.997} \\
\ours{} & \textbf{.998} & \textbf{.996} & .973 & .962 \\
\bottomrule
\end{tabular}

\vspace{1pt}
\setlength{\tabcolsep}{1.5pt}
\begin{tabular}[b]{lcccccc}
\toprule
\multicolumn{7}{l}{\emph{(b) Flow generation}} \\
& \multicolumn{4}{c}{Vertex flow} & \multicolumn{2}{c}{Edge flow} \\
\cmidrule(lr){2-5}\cmidrule(lr){6-7}
& \multicolumn{2}{c}{\cdo{}$\downarrow$} & \multicolumn{2}{c}{\hd{}$\downarrow$} & \multicolumn{2}{c}{F1$\uparrow$} \\
\cmidrule(lr){2-3}\cmidrule(lr){4-5}\cmidrule(lr){6-7}
Method & Toys4K & Obj. & Toys4K & Obj. & Toys4K & Obj. \\
\midrule
MeshFlow & .0260 & .0298 & .144 & .172 & -- & -- \\
LATO.2 & .0149 & .0145 & .117 & .144 & .729 & .673 \\
\ours{} & \textbf{.0091} & \textbf{.0079} & \textbf{.106} & \textbf{.111} & \textbf{.795} & \textbf{.820} \\
\bottomrule
\end{tabular}
}%
\setbox\tabBbox\vbox{\hsize=0.56\textwidth\@parboxrestore
\caption{\textbf{Ablations.} (a) EdgeVAE variants scored on 5,000 validation samples with the all-pair edge metric, our model first and one component changed per row. Precision and recall are computed over all vertex pairs of a mesh and averaged over meshes. (b) Anchor generation, the anchor generated by a flow against our anchor generator, the two sharing the same downstream stages on Objaverse.}
\label{tab:ablation}
\vfill
\centering
\scriptsize
\renewcommand{\arraystretch}{0.92}
\newcommand{\vm}[1]{\multicolumn{4}{c}{#1}}
\newcommand{\fm}[1]{\multicolumn{3}{c}{#1}}
\begin{tabular}[b]{l@{\hspace{2pt}}*{12}{@{}>{\centering\arraybackslash}p{0.11in}@{}}}
\toprule
\multicolumn{13}{l}{\emph{(a) EdgeVAE variants}} \\
Variant & \vm{precision} & \vm{recall} & \vm{F1} \\
\midrule
\ours{} & \vm{\textbf{.95180}} & \vm{\textbf{.99019}} & \vm{\textbf{.96965}} \\
mean pooling and broadcast & \vm{.81978} & \vm{.96361} & \vm{.88111} \\
w/o vertex-link encoding & \vm{.61701} & \vm{.85324} & \vm{.70793} \\
slotted vertex neighbor encoding & \vm{.93806} & \vm{.98959} & \vm{.96167} \\
w/o RoPE & \vm{.36967} & \vm{.50487} & \vm{.40872} \\
joint latent for vertices and edges & \vm{.90258} & \vm{.95743} & \vm{.92743} \\
\midrule
\multicolumn{13}{l}{\emph{(b) Anchor generation}} \\
Anchor & \fm{\cdl{}$\downarrow$} & \fm{\cdo{}$\downarrow$} & \fm{\hd{}$\downarrow$} & \fm{\nc{}$\uparrow$} \\
\midrule
generated by a flow & \fm{.01007} & \fm{.01422} & \fm{.0971} & \fm{.8012} \\
anchor generator (\ours{}) & \fm{\textbf{.00649}} & \fm{\textbf{.00933}} & \fm{\textbf{.0789}} & \fm{\textbf{.8529}} \\
\bottomrule
\end{tabular}
}%
\setlength{\tabH}{\ht\tabAbox}\ifdim\ht\tabBbox>\tabH\setlength{\tabH}{\ht\tabBbox}\fi
\noindent\vbox to\tabH{\unvbox\tabAbox}\hfill\vbox to\tabH{\unvbox\tabBbox}
\makeatother
\end{table}

\subsection{Empirical Study}
\label{sec:ablations}

\textbf{Qualitative results.} In Figure~\ref{fig:qual}, \ours{} follows the reference surface closely and places vertices where the artist did, while the autoregressive baselines leave parts of the shape uncovered. MeshFlow, in comparison, generates meshes with an overly low face count that cannot describe the object shape precisely, which explains its low \hd{} yet higher \cdl{} and \cdo{}. LATO.2 also has difficulty capturing fine details and does not handle thin structures well. More examples are in Appendix~\ref{app:qual_more} and the limitations are discussed in Appendix~\ref{app:limit}.

\textbf{Ablations.} Table~\ref{tab:ablation} ablates the critical components of our VAE backbone. We replace the spatial-aware compression with naive mean pooling and broadcast, remove vertex-link encoding, swap vertex-link encoding with slotted vertex neighbor encoding, and remove RoPE in the sparse transformer blocks. All four variants degrade the performance, justifying our VAE design. Removing RoPE causes the largest drop, since the connections between vertex voxels depend on their relative positions. The slotted vertex neighbor encoding comes closest to \ours{}. It is trained from scratch for the full schedule, so this small gap is not an artifact of a shorter schedule. In every variant the precision drops more than the recall, so the changed components mainly prevent connections that the artist did not draw. We also show the results when the same VAE backbone is used to encode vertex position and edge connection simultaneously. The degradation is noticeable and is amplified by inaccurate vertex subdivision and pruning, because the error accumulates fast during inference. We also ablate the use of the anchor generator versus a flow matching network. The anchor generator shows significant advantages over a flow matching network in obtaining the latent voxel positions. Both anchors feed the same vertex and edge flows, so the gap in Table~\ref{tab:ablation}(b) comes from the anchor alone. Notably, the anchor generator is based on a VAE backbone and therefore also generates with diversity, as shown in Figure~\ref{fig:s1_seeds} of the supplementary.

\textbf{Limitations and future work.} While \ours{} demonstrates its effectiveness in our evaluation, we notice some limitations imposed by the network. Specifically, for complex meshes over 10K vertices, the $512^3$ resolution might not be sufficient and fine details might not be generated faithfully. We also find that the flow matching networks trained with teacher forcing encounter a domain gap when cascaded, and we believe further adaptation will unlock more of the edge flow's potential.

\section{Conclusion}
\label{sec:conclusion}
We presented \ours{}, which generates artisan meshes by flow matching entirely in spatially compressed latents. The pipeline samples the coarse vertex occupancy with a VAE in one forward pass and then generates vertex positions and edge connections in turn. The latents are compact because of the VAE design. One hierarchical sparse transformer backbone is anchored to the vertex voxels and compressed spatially through a learned projection. It is instantiated separately for vertex positions and for edge connections, and the connections reach it as the vertex-link encoding. At $0.13$ tokens per face, \ours{} shows advantages over previous state of the art on Objaverse and generalizes to Toys4K.

\bibliography{refs,arxiv_extra}

\begin{thebibliography}{36}
\providecommand{\natexlab}[1]{#1}
\providecommand{\url}[1]{\texttt{#1}}
\expandafter\ifx\csname urlstyle\endcsname\relax
  \providecommand{\doi}[1]{doi: #1}\else
  \providecommand{\doi}{doi: \begingroup \urlstyle{rm}\Url}\fi

\bibitem[Alliegro et~al.(2023)Alliegro, Siddiqui, Tommasi, and Nie{\ss}ner]{polydiff}
Antonio Alliegro, Yawar Siddiqui, Tatiana Tommasi, and Matthias Nie{\ss}ner.
\newblock {PolyDiff}: Generating 3d polygonal meshes with diffusion models.
\newblock \emph{arXiv preprint arXiv:2312.11417}, 2023.

\bibitem[Chen et~al.(2025{\natexlab{a}})Chen, Cai, Chen, Xie, Yang, Tang, Li, Lu, and Han]{dcae}
Junyu Chen, Han Cai, Junsong Chen, Enze Xie, Shang Yang, Haotian Tang, Muyang Li, Yao Lu, and Song Han.
\newblock Deep compression autoencoder for efficient high-resolution diffusion models.
\newblock In \emph{ICLR}, 2025{\natexlab{a}}.
\newblock arXiv:2410.10733.

\bibitem[Chen et~al.(2025{\natexlab{b}})Chen, He, Huang, Ye, Chen, Tang, Chen, Cai, Yang, Yu, Lin, and Zhang]{meshanything}
Yiwen Chen, Tong He, Di~Huang, Weicai Ye, Sijin Chen, Jiaxiang Tang, Xin Chen, Zhongang Cai, Lei Yang, Gang Yu, Guosheng Lin, and Chi Zhang.
\newblock {MeshAnything}: Artist-created mesh generation with autoregressive transformers.
\newblock In \emph{ICLR}, 2025{\natexlab{b}}.

\bibitem[Chen et~al.(2025{\natexlab{c}})Chen, Wang, Luo, Wang, Chen, Zhu, Zhang, and Lin]{meshanythingv2}
Yiwen Chen, Yikai Wang, Yihao Luo, Zhengyi Wang, Zilong Chen, Jun Zhu, Chi Zhang, and Guosheng Lin.
\newblock {MeshAnything V2}: Artist-created mesh generation with adjacent mesh tokenization.
\newblock In \emph{ICCV}, 2025{\natexlab{c}}.

\bibitem[Daunhawer et~al.(2022)Daunhawer, Sutter, Chin-Cheong, Palumbo, and Vogt]{daunhawer2022limitations}
Imant Daunhawer, Thomas~M. Sutter, Kieran Chin-Cheong, Emanuele Palumbo, and Julia~E. Vogt.
\newblock On the limitations of multimodal {VAEs}.
\newblock In \emph{ICLR}, 2022.

\bibitem[Deitke et~al.(2023)Deitke, Schwenk, Salvador, Weihs, Michel, VanderBilt, Schmidt, Ehsani, Kembhavi, and Farhadi]{objaverse}
Matt Deitke, Dustin Schwenk, Jordi Salvador, Luca Weihs, Oscar Michel, Eli VanderBilt, Ludwig Schmidt, Kiana Ehsani, Aniruddha Kembhavi, and Ali Farhadi.
\newblock {Objaverse}: A universe of annotated 3{D} objects.
\newblock In \emph{CVPR}, 2023.

\bibitem[He et~al.(2025)He, Chen, Huang, Liu, Huang, Ouyang, Yuan, and Li]{meshcraft}
Xianglong He, Junyi Chen, Di~Huang, Zexiang Liu, Xiaoshui Huang, Wanli Ouyang, Chun Yuan, and Yangguang Li.
\newblock {MeshCraft}: Exploring efficient and controllable mesh generation with flow-based {DiTs}.
\newblock \emph{arXiv preprint arXiv:2503.23022}, 2025.

\bibitem[Javaloy et~al.(2022)Javaloy, Meghdadi, and Valera]{javaloy2022modality}
Adri{\'a}n Javaloy, Maryam Meghdadi, and Isabel Valera.
\newblock Mitigating modality collapse in multimodal {VAEs} via impartial optimization.
\newblock In \emph{ICML}, 2022.

\bibitem[Kim et~al.(2026)Kim, Lan, Fortes, Chen, and Pan]{fastmesh}
Jeonghwan Kim, Yushi Lan, Armando Fortes, Yongwei Chen, and Xingang Pan.
\newblock {FastMesh}: Efficient artistic mesh generation via component decoupling.
\newblock In \emph{International Conference on 3D Vision (3DV)}, 2026.
\newblock arXiv:2508.19188.

\bibitem[Kingma \& Welling(2014)Kingma and Welling]{vae}
Diederik~P. Kingma and Max Welling.
\newblock Auto-encoding variational {B}ayes.
\newblock In \emph{ICLR}, 2014.
\newblock arXiv:1312.6114.

\bibitem[Li et~al.(2026{\natexlab{a}})Li, Erkoç, Sirigatti, Rosov, Li, Dai, and Nießner]{triflow}
Haoxuan Li, Ziya Erkoç, Daniele Sirigatti, Vladislav Rosov, Lei Li, Angela Dai, and Matthias Nießner.
\newblock {TriFlow}: Generating artist-like 3d mesh topology via nearest-vertex vector fields.
\newblock In \emph{ECCV}, 2026{\natexlab{a}}.

\bibitem[Li et~al.(2026{\natexlab{b}})Li, Toisoul, Monnier, Shapovalov, Ranjan, Tan, and Vedaldi]{meshflow}
Weiyu Li, Antoine Toisoul, Tom Monnier, Roman Shapovalov, Rakesh Ranjan, Ping Tan, and Andrea Vedaldi.
\newblock {MeshFlow}: Efficient artistic mesh generation via {MeshVAE} and flow-based diffusion transformer.
\newblock In \emph{CVPR}, 2026{\natexlab{b}}.

\bibitem[Li et~al.(2025)Li, Wang, Zheng, Luo, and Wen]{sparc3d}
Zhihao Li, Yufei Wang, Heliang Zheng, Yihao Luo, and Bihan Wen.
\newblock {Sparc3D}: Sparse representation and construction for high-resolution {3D} shapes modeling.
\newblock In \emph{NeurIPS}, 2025.

\bibitem[Lionar et~al.(2025)Lionar, Liang, and Lee]{treemeshgpt}
Stefan Lionar, Jiabin Liang, and Gim~Hee Lee.
\newblock {TreeMeshGPT}: Artistic mesh generation with autoregressive tree sequencing.
\newblock In \emph{CVPR}, 2025.

\bibitem[Lipman et~al.(2023)Lipman, Chen, Ben-Hamu, Nickel, and Le]{flowmatching}
Yaron Lipman, Ricky T.~Q. Chen, Heli Ben-Hamu, Maximilian Nickel, and Matt Le.
\newblock Flow matching for generative modeling.
\newblock In \emph{ICLR}, 2023.

\bibitem[Liu et~al.(2023)Liu, Gong, and Liu]{rectifiedflow}
Xingchao Liu, Chengyue Gong, and Qiang Liu.
\newblock Flow straight and fast: Learning to generate and transfer data with rectified flow.
\newblock In \emph{ICLR}, 2023.

\bibitem[Long et~al.(2026)Long, Zhao, Lin, Zhang, Guo, Liang, Xu, Hladk{\'y}, Nie{\ss}ner, Hu, and Yang]{lato2}
Hang Long, Tianhao Zhao, Junkai Lin, Youjia Zhang, Huipeng Guo, Rendong Liang, Jiale Xu, Jozef Hladk{\'y}, Matthias Nie{\ss}ner, Yuanming Hu, and Wei Yang.
\newblock {LATO.2}: Factorized 3d mesh generation with vertex and topology flow.
\newblock \emph{arXiv preprint arXiv:2607.10623}, 2026.

\bibitem[Nash et~al.(2020)Nash, Ganin, Eslami, and Battaglia]{polygen}
Charlie Nash, Yaroslav Ganin, S.~M.~Ali Eslami, and Peter~W. Battaglia.
\newblock {PolyGen}: An autoregressive generative model of 3{D} meshes.
\newblock In \emph{ICML}, 2020.
\newblock arXiv:2002.10880.

\bibitem[Palumbo et~al.(2023)Palumbo, Daunhawer, and Vogt]{palumbo2023mmvaeplus}
Emanuele Palumbo, Imant Daunhawer, and Julia~E. Vogt.
\newblock {MMVAE+}: Enhancing the generative quality of multimodal {VAEs} without compromises.
\newblock In \emph{ICLR}, 2023.

\bibitem[Rahimi \& Recht(2007)Rahimi and Recht]{rff}
Ali Rahimi and Benjamin Recht.
\newblock Random features for large-scale kernel machines.
\newblock In \emph{NeurIPS}, 2007.

\bibitem[Ridnik et~al.(2021)Ridnik, Ben-Baruch, Zamir, Noy, Friedman, Protter, and Zelnik-Manor]{asl}
Tal Ridnik, Emanuel Ben-Baruch, Nadav Zamir, Asaf Noy, Itamar Friedman, Matan Protter, and Lihi Zelnik-Manor.
\newblock Asymmetric loss for multi-label classification.
\newblock In \emph{ICCV}, 2021.

\bibitem[Rombach et~al.(2022)Rombach, Blattmann, Lorenz, Esser, and Ommer]{ldm}
Robin Rombach, Andreas Blattmann, Dominik Lorenz, Patrick Esser, and Bj{\"o}rn Ommer.
\newblock High-resolution image synthesis with latent diffusion models.
\newblock In \emph{CVPR}, 2022.
\newblock arXiv:2112.10752.

\bibitem[Shen et~al.(2024)Shen, Li, Law, Atzmon, Fidler, Lucas, Gao, and Sharp]{spacemesh}
Tianchang Shen, Zhaoshuo Li, Marc Law, Matan Atzmon, Sanja Fidler, James Lucas, Jun Gao, and Nicholas Sharp.
\newblock {SpaceMesh}: A continuous representation for learning manifold surface meshes.
\newblock In \emph{SIGGRAPH Asia}, 2024.

\bibitem[Siddiqui et~al.(2024)Siddiqui, Alliegro, Artemov, Tommasi, Sirigatti, Levy, Dai, and Nie{\ss}ner]{meshgpt}
Yawar Siddiqui, Antonio Alliegro, Alexey Artemov, Tatiana Tommasi, Daniele Sirigatti, Vladislav Levy, Angela Dai, and Matthias Nie{\ss}ner.
\newblock {MeshGPT}: Generating triangle meshes with decoder-only transformers.
\newblock In \emph{CVPR}, 2024.
\newblock arXiv:2311.15475.

\bibitem[Stojanov et~al.(2021)Stojanov, Thai, and Rehg]{toys4k}
Stefan Stojanov, Anh Thai, and James~M. Rehg.
\newblock Using shape to categorize: Low-shot learning with an explicit shape bias.
\newblock In \emph{CVPR}, 2021.

\bibitem[Tancik et~al.(2020)Tancik, Srinivasan, Mildenhall, Fridovich-Keil, Raghavan, Singhal, Ramamoorthi, Barron, and Ng]{tancik}
Matthew Tancik, Pratul~P. Srinivasan, Ben Mildenhall, Sara Fridovich-Keil, Nithin Raghavan, Utkarsh Singhal, Ravi Ramamoorthi, Jonathan~T. Barron, and Ren Ng.
\newblock Fourier features let networks learn high frequency functions in low dimensional domains.
\newblock In \emph{NeurIPS}, 2020.

\bibitem[Tang et~al.(2025)Tang, Li, Hao, Liu, Zeng, Liu, and Zhang]{edgerunner}
Jiaxiang Tang, Zhaoshuo Li, Zekun Hao, Xian Liu, Gang Zeng, Ming-Yu Liu, and Qinsheng Zhang.
\newblock {EdgeRunner}: Auto-regressive auto-encoder for artistic mesh generation.
\newblock In \emph{ICLR}, 2025.

\bibitem[Wang et~al.(2026{\natexlab{a}})Wang, Guo, Liu, Zou, Zhang, Quan, Liang, Cao, and Yan]{face}
Hanxiao Wang, Yuan-Chen Guo, Ying-Tian Liu, Zi-Xin Zou, Biao Zhang, Weize Quan, Ding Liang, Yan-Pei Cao, and Dong-Ming Yan.
\newblock {FACE}: A face-based autoregressive representation for high-fidelity and efficient mesh generation.
\newblock In \emph{CVPR}, 2026{\natexlab{a}}.

\bibitem[Wang et~al.(2026{\natexlab{b}})Wang, Liu, Guo, Feng, Zou, Liang, Zhang, and Cao]{nexus}
Hanxiao Wang, Ying-Tian Liu, Yuan-Chen Guo, Qi-Yuan Feng, Zi-Xin Zou, Ding Liang, Biao Zhang, and Yan-Pei Cao.
\newblock {Nexus}: Native mesh generation with diffusion.
\newblock \emph{ACM Transactions on Graphics}, 45\penalty0 (4), 2026{\natexlab{b}}.

\bibitem[Weng et~al.(2025)Weng, Zhao, Lei, Yang, Liu, Lai, Chen, Liu, Jiang, Guo, Zhang, Gao, and Chen]{bpt}
Haohan Weng, Zibo Zhao, Biwen Lei, Xianghui Yang, Jian Liu, Zeqiang Lai, Zhuo Chen, Yuhong Liu, Jie Jiang, Chunchao Guo, Tong Zhang, Shenghua Gao, and C.~L.~Philip Chen.
\newblock Scaling mesh generation via compressive tokenization.
\newblock In \emph{CVPR}, 2025.

\bibitem[Xiang et~al.(2025)Xiang, Lv, Xu, Deng, Wang, Zhang, Chen, Tong, and Yang]{trellis}
Jianfeng Xiang, Zelong Lv, Sicheng Xu, Yu~Deng, Ruicheng Wang, Bowen Zhang, Dong Chen, Xin Tong, and Jiaolong Yang.
\newblock Structured 3{D} latents for scalable and versatile 3{D} generation.
\newblock In \emph{CVPR}, 2025.

\bibitem[Xiang et~al.(2026)Xiang, Chen, Xu, Wang, Lv, Deng, Zhu, Dong, Zhao, Yuan, and Yang]{trellis2}
Jianfeng Xiang, Xiaoxue Chen, Sicheng Xu, Ruicheng Wang, Zelong Lv, Yu~Deng, Hongyuan Zhu, Yue Dong, Hao Zhao, Nicholas~Jing Yuan, and Jiaolong Yang.
\newblock Native and compact structured latents for 3d generation.
\newblock In \emph{CVPR}, 2026.

\bibitem[Xu et~al.(2026)Xu, Xue, Dong, Wan, Zhu, Li, Dou, Lin, Xin, Liu, Wang, and Komura]{meshmosaic}
Rui Xu, Tianyang Xue, Qiujie Dong, Le~Wan, Zhe Zhu, Peng Li, Zhiyang Dou, Cheng Lin, Shiqing Xin, Yuan Liu, Wenping Wang, and Taku Komura.
\newblock {MeshMosaic}: Scaling artist mesh generation via local-to-global assembly.
\newblock In \emph{CVPR}, 2026.

\bibitem[Zhao et~al.(2025)Zhao, Ye, Wang, Liu, Chen, Wang, and Zhu]{deepmesh}
Ruowen Zhao, Junliang Ye, Zhengyi Wang, Guangce Liu, Yiwen Chen, Yikai Wang, and Jun Zhu.
\newblock {DeepMesh}: Auto-regressive artist-mesh creation with reinforcement learning.
\newblock In \emph{ICCV}, 2025.

\bibitem[Zhao et~al.(2026)Zhao, Zhang, Long, Zhang, Li, Yang, Zhang, Hladk{\'y}, Nie{\ss}ner, and Yang]{lato1}
Tianhao Zhao, Youjia Zhang, Hang Long, Jinshen Zhang, Wenbing Li, Yang Yang, Gongbo Zhang, Jozef Hladk{\'y}, Matthias Nie{\ss}ner, and Wei Yang.
\newblock {LATO}: 3d mesh flow matching with structured topology preserving latents.
\newblock In \emph{ICML}, 2026.

\bibitem[Zhao et~al.(2023)Zhao, Liu, Chen, Zeng, Wang, Cheng, Fu, Chen, Yu, and Gao]{michelangelo}
Zibo Zhao, Wen Liu, Xin Chen, Xianfang Zeng, Rui Wang, Pei Cheng, Bin Fu, Tao Chen, Gang Yu, and Shenghua Gao.
\newblock {Michelangelo}: Conditional 3{D} shape generation based on shape-image-text aligned latent representation.
\newblock In \emph{NeurIPS}, 2023.

\end{thebibliography}
\bibliographystyle{preprint}

\appendix
\section{Objectives and Flow Matching Details}
\label{app:flow}
Both stages of \ours{} follow the latent-diffusion recipe~\citep{ldm}: a VAE~\citep{vae} maps a mesh to a compact latent, and a flow model is trained in that latent space. The VAE pairs an encoder $q_\phi(\mathbf{z}\mid\mathbf{x})$, which maps an input $\mathbf{x}$ to a distribution over latents $\mathbf{z}$, with a decoder that reconstructs $\mathbf{x}$ from $\mathbf{z}$. The two are trained jointly by minimizing
\begin{equation}
\mathcal{L}=\mathcal{L}_{\mathrm{rec}}+\lambda\,\mathrm{KL}\big(q_\phi(\mathbf{z}\mid\mathbf{x})\,\|\,\mathcal{N}(\mathbf{0},\mathbf{I})\big),\qquad \mathbf{z}\sim q_\phi(\mathbf{z}\mid\mathbf{x}),
\end{equation}
where the reconstruction loss $\mathcal{L}_{\mathrm{rec}}$ measures how well the decoder recovers $\mathbf{x}$ from $\mathbf{z}$, the KL term keeps the encoded latents close to a standard Gaussian prior, and $\lambda$ weights the two, with $\lambda=10^{-3}$ for VertexVAE and EdgeVAE. In our case $\mathbf{x}$ is the mesh $(\mathbf{V},\mathbf{A})$ and $\mathbf{z}$ its latent voxels. Sections 3.2 and 3.3 of the main paper give the reconstruction term of each VAE.

Both latents are generated with rectified flow~\citep{rectifiedflow,flowmatching}: given a latent $\mathbf{z}_0$ and Gaussian noise $\boldsymbol{\epsilon}$, the interpolant $\mathbf{z}_t=(1-t)\,\mathbf{z}_0+t\,\boldsymbol{\epsilon}$ is formed for $t\in[0,1]$ and a network $\mathbf{u}_\theta$ predicts its velocity,
\begin{equation}
\mathcal{L}_{\mathrm{flow}}=\mathbb{E}_{t,\mathbf{z}_0,\boldsymbol{\epsilon}}\big\|\mathbf{u}_\theta(\mathbf{z}_t,t,C)-(\boldsymbol{\epsilon}-\mathbf{z}_0)\big\|^{2},
\label{eq:flow}
\end{equation}
where $C$ is the condition of the flow. At inference, $\mathbf{z}_t$ is integrated from noise at $t=1$ to $t=0$ with classifier-free guidance and decoded by the frozen decoder. Both flows use the sparse transformer of~\citet{trellis,trellis2}, with one token per latent voxel, and differ from it only in the condition: the dense point cloud enters through the tokens of a frozen shape encoder~\citep{michelangelo} by cross-attention. The vertex flow generates the 16-channel VertexVAE latent on the generated anchor and takes the vertex budget $N$ through adaptive layer normalization. The edge flow generates the 16-channel EdgeVAE latent on the same anchor and takes the vertex latent of the same voxel by concatenation. Apart from their size, only the latent space and the condition differ between the two flows.

\section{Implementation Details}
\label{app:impl}
The anchor generator is a sparse VAE of convolution and attention blocks with 613M parameters and a KL weight of $10^{-4}$, trained on meshes with at most 20K vertex voxels at $512^3$. It reads the input surface in two forms. One is its occupancy rasterized at $64^3$ and $32^3$. The other is a set of area-weighted point samples with normals, drawn from a flexible dual-grid remesh of the surface at $512^3$. This is the representation a shape-generation pipeline emits, so the anchor is decided from a tessellation-free surface. The flows are conditioned on the point cloud through the tokens of the frozen shape encoder. All models are trained with Adam. VertexVAE and EdgeVAE are trained for 270K and 300K steps at a learning rate of 1e-4 and a batch size of 96. The anchor generator is trained for 27K steps at a learning rate of 5e-5 and a batch size of 64. The vertex flow is trained for 90K steps at a batch size of 512 and the edge flow for 110K steps at a batch size of 480, both at a learning rate of 1e-4 with an exponential moving average of the weights. At inference the anchor generator decodes its posterior mean in a single forward pass, the setting of Table 2 of the main paper; Appendix~\ref{app:s1_vae} compares it with a posterior sample. The vertex flow (1.27B) is sampled for 50 steps with compositional classifier-free guidance over VertexVAE (429M). The edge flow (2.37B) is sampled for 12 steps with guidance scale 3.0 over EdgeVAE (483M), followed by the connection head over all vertex pairs and 3-cycle face enumeration. Every 3-cycle of the predicted edge graph is emitted as a face, with no face classifier and no manifoldness constraint. 

\section{Evaluation Protocol}
\label{app:protocol}
Each baseline is run from its released weights under its stock inference setting, fed the condition it was designed for. The autoregressive methods stop at their own face ceilings rather than at a budget we impose. MeshAnything, MeshAnything V2 and BPT are therefore scored on the evaluation meshes whose reference face count is inside their published caps of 800, 1,600 and 8,000. This is the range each method was built for. DeepMesh, TreeMeshGPT and FastMesh publish no cap and are scored on the full sets. Scoring the capped methods on the full sets instead worsens every one of their numbers and widens every gap of Table 2 of the main paper. Table~\ref{tab:arcap} gives that comparison together with \ours{} restricted to the same meshes. The topology flow of LATO.2 accepts a bounded number of input vertices, and no setting lifts the bound. The comparison given reference vertices therefore runs on the 147 Objaverse and 465 Toys4K meshes it admits. \ours{} is scored on the same meshes. The vertex count is a user-specified knob in recent methods~\citep{lato2,meshcraft,meshanythingv2,edgerunner}. \ours{} and LATO.2 receive the reference vertex count, which matches the deployment interface of LATO.2. The remaining baselines expose no working count input. The released \texttt{--num\_verts} of MeshFlow has no effect. They generate to their own termination, at their own face count (Table~\ref{tab:incidence_full}). DeepMesh outputs are axis-permuted as required by its release, and TreeMeshGPT inputs are decimated per its documentation. Prediction and reference are each normalized to the $[-0.5,0.5]$ box per mesh before sampling 50K surface points, and the metrics are computed between the two point sets.

\begin{table}[t]
\centering
\caption{Vertex and face budgets relative to the reference on Toys4K, means per mesh. Bold: closest to the reference among the flow-based methods. Underlined: among the autoregressive methods.}
\label{tab:incidence_full}
\small
\setlength{\tabcolsep}{4pt}
\begin{tabular}{lccc}
\toprule
Method & vertices / ref. & faces / ref. & faces / vertex \\
\midrule
Reference & 1.00 & 1.00 & 1.48 \\
\midrule
MeshAnything~\citep{meshanything} & 0.24 & 0.33 & \underline{1.74} \\
MeshAnythingV2~\citep{meshanythingv2} & 0.74 & \underline{1.01} & 1.89 \\
TreeMeshGPT~\citep{treemeshgpt} & 0.58 & 0.78 & 1.88 \\
BPT~\citep{bpt} & 0.43 & 0.61 & 1.96 \\
DeepMesh~\citep{deepmesh} & \underline{1.03} & 1.47 & 1.90 \\
FastMesh~\citep{fastmesh} & 1.30 & 6.83 & 7.67 \\
\midrule
MeshFlow~\citep{meshflow} & 0.35 & 0.58 & 2.23 \\
LATO.2~\citep{lato2} & 0.76 & 1.58 & 2.92 \\
\ours{} (ours) & \textbf{1.06} & \textbf{1.34} & \textbf{1.92} \\
\bottomrule
\end{tabular}
\end{table}

VAE round trips in Table 3 of the main paper decode the posterior mean, except the vertex VAE of LATO.2 and the joint VAE of LATO, which are run with their default posterior sampling. The VAE of MeshFlow and the topology stage of LATO.2 are scored on the meshes below their vertex limits. The vertex VAE of LATO.2 is scored on its decoded voxel coordinates before its sub-voxel offset and that of MeshFlow on its regressed positions quantized to $512^3$. MeshFlow generates vertices and edges jointly and has no edge flow given vertices. The joint VAE of LATO decodes its vertex voxels freely for the vertex column and is teacher-forced to the reference vertex voxels for the edge column, with its released connection head at its default threshold. The edge-flow F1 compares the generated edges with the artist's edges over the reference vertex voxels at $512^3$. The vertices of LATO.2 ride its own $1024^3$ quantization, so each is snapped to the nearest reference vertex voxel within one cell before its edges are read, which favors it. Each variant in Table 4 of the main paper changes one component of \ours{} and is evaluated on the same 5,000 validation samples with the same all-pair edge metric, with precision and recall computed over all vertex pairs of a mesh and averaged over meshes. The slotted vertex neighbor encoding, the joint latent for vertices and edges, and the flow-generated anchor are trained from scratch for the same number of steps as the deployed model they replace, so that their smaller gaps are not an artifact of the schedule. The remaining variants are finetuned from the deployed VAE for 25,000 steps, until their performance stabilizes, because training every variant from scratch is costly. Their gaps are large and come from a missing component that the model cannot recover, so training them from scratch is unnecessary.

\section{Structural Statistics of the Generated Meshes}
\label{app:struct}
Surface metrics do not see connectivity. Table~\ref{tab:struct} therefore reports statistics of the generated graphs themselves. The end-to-end methods are scored on the full evaluation sets. The block given the reference vertices is scored on the meshes the topology flow of LATO.2 admits. An edge is boundary, manifold or non-manifold according to whether one, two or more faces are incident on it. The three shares are computed per mesh and averaged. The artisan meshes are manifold almost everywhere, with 23\% boundary edges on Toys4K from open sheets. Valence is the mean vertex degree, and $W_1$ the earth mover's distance between the mesh's valence histogram and that of its reference. Non-manifold edges are the defect of every method that decides connectivity by classification rather than by serializing faces. The autoregressive methods except FastMesh stay below 7\% and are the closest to the artisan meshes on every statistic except the boundary share. MeshFlow, LATO.2 and \ours{} over-connect. End to end, \ours{} is close to MeshFlow, below it on Objaverse and above it on Toys4K. Both are well ahead of LATO.2. The MeshFlow figures come with the budgets of Table~\ref{tab:incidence_full}, a mesh with 0.35 of the reference vertices and 0.58 of its faces, so they describe a coarser mesh than the reference rather than the reference's connectivity. On the reference vertices, the non-manifold share of \ours{} is roughly half of that of the LATO.2 topology flow, and its valence histogram is the closest to the artist's. This locates the end-to-end defect in the vertices the edge stage is run on rather than in the stage itself, the limitation named in Appendix~\ref{app:limit}. On boundary edges \ours{} is the closest to the artisan meshes on Toys4K and overshoots on Objaverse. Triangle-shape statistics (minimum angle, sliver fraction, edge-length spread) are omitted, as they separate no method from the artisan meshes, which themselves contain many sliver faces.

\begin{table}[h]
\centering
\caption{Structural statistics of the generated meshes, means per mesh. B / M / NM: boundary, manifold and non-manifold edge shares in \%. Val.: mean vertex valence. $W_1$: distance of the valence histogram to the reference's (lower is closer).}
\label{tab:struct}
\footnotesize
\setlength{\tabcolsep}{3pt}
\begin{tabular}{lccccc@{\hspace{8pt}}ccccc}
\toprule
& \multicolumn{5}{c}{Objaverse} & \multicolumn{5}{c}{Toys4K} \\
\cmidrule(lr){2-6}\cmidrule(lr){7-11}
Method & B & M & NM & Val. & $W_1$ & B & M & NM & Val. & $W_1$ \\
\midrule
Reference & 5.6 & 94.4 & 0.0 & 5.52 & 0.00 & 23.0 & 76.8 & 0.2 & 4.84 & 0.00 \\
\midrule
MeshAnything~\citep{meshanything} & 11.3 & 86.3 & 2.4 & 5.04 & 0.87 & 10.6 & 85.7 & 3.7 & 5.33 & 1.20 \\
MeshAnythingV2~\citep{meshanythingv2} & 10.5 & 83.0 & 6.5 & 5.40 & 0.89 & 12.3 & 81.8 & 5.9 & 5.71 & 1.37 \\
TreeMeshGPT~\citep{treemeshgpt} & 9.2 & 89.7 & 1.1 & 5.42 & 0.61 & 4.4 & 95.1 & 0.6 & 5.74 & 1.17 \\
BPT~\citep{bpt} & 9.6 & 84.2 & 6.2 & 5.72 & 0.87 & 5.3 & 92.1 & 2.6 & 5.86 & 1.28 \\
DeepMesh~\citep{deepmesh} & 10.6 & 86.1 & 3.3 & 5.55 & 0.72 & 7.5 & 90.5 & 2.0 & 5.78 & 1.24 \\
FastMesh~\citep{fastmesh} & 0.8 & 11.8 & 87.4 & 11.50 & 5.58 & 0.4 & 19.8 & 79.8 & 9.98 & 4.96 \\
\midrule
MeshFlow~\citep{meshflow} & 5.4 & 65.5 & 29.1 & 6.04 & 1.06 & 2.3 & 82.2 & 15.5 & 5.99 & 1.36 \\
LATO.2~\citep{lato2} & 6.6 & 49.4 & 43.9 & 6.90 & 1.76 & 3.4 & 62.2 & 34.4 & 6.73 & 2.01 \\
\ours{} (ours) & 16.8 & 55.4 & 27.7 & 5.68 & 1.11 & 17.2 & 65.3 & 17.5 & 5.55 & 1.38 \\
\midrule
\multicolumn{11}{l}{\emph{given the reference vertices}} \\
LATO.2 T-Flow~\citep{lato2} & 4.7 & 62.9 & 32.4 & 6.16 & 0.95 & 2.5 & 76.4 & 21.1 & 6.20 & 1.52 \\
\ours{} edge flow & 4.8 & 80.3 & 14.9 & 5.87 & 0.67 & 4.6 & 83.7 & 11.7 & 5.92 & 1.38 \\
\bottomrule
\end{tabular}
\end{table}

\section{Token Budget against Mesh Size}
\label{app:tokens}
\begin{figure}[t]
\centering
\includegraphics[width=\linewidth]{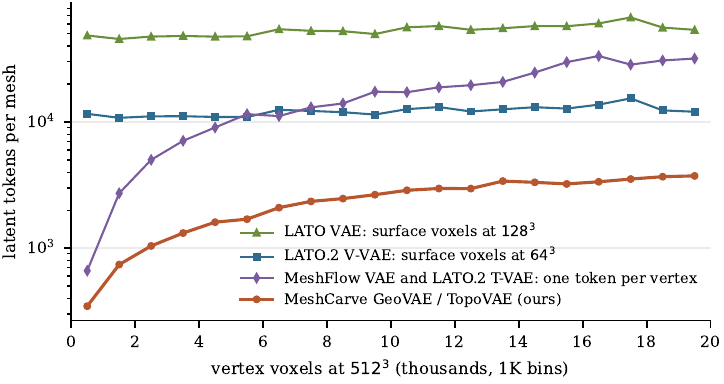}
\caption{Latent sequence length of each VAE against mesh size on 5,000 random training meshes, averaged within 1K-wide bins of the vertex-voxel count at $512^3$. Every length is determined by the ground-truth mesh alone, so the figure isolates compression from generation. Surface voxels are those hit by a dense sample of the surface.}
\label{fig:tokens}
\end{figure}

Table 1 of the main paper pools the token budget over the evaluation sets. Figure~\ref{fig:tokens} decomposes it by mesh size on 5,000 meshes drawn at random from the training set, for the VAEs alone. The meshes are binned by their number of vertex voxels at $512^3$. Each curve is the latent sequence length one VAE assigns to the ground-truth mesh, which follows from the mesh alone. For \ours{} it is the number of vertex voxels at $64^3$. For LATO it is the number of surface voxels at $128^3$, and for the vertex VAE of LATO.2 at $64^3$. For MeshFlow and the topology VAE of LATO.2 it is one token per vertex. The figure therefore measures compression only, with no generation involved. The two surface-anchored latents pay for the surface rather than the mesh, and stay flat as the vertex count grows. The per-vertex latents grow with it. Our latent is the shortest on every bin, and the distance widens with mesh size, as spatial-aware compression packs more vertices into each token.

\section{Autoregressive Baselines outside Their Face Caps}
\label{app:arcap}
Table 2 of the main paper scores MeshAnything, MeshAnything V2 and BPT inside their published face caps, which is the range each method was built for. Table~\ref{tab:arcap} repeats the comparison on the full evaluation sets. These include the meshes past the cap, on which each method truncates or stops early. Every number of every method is worse on the full sets than inside its cap, so the restriction of Table 2 of the main paper is the setting that favors them. Each block also gives \ours{} on exactly the meshes inside the cap, and the last row gives \ours{} on the full sets, as in Table 2 of the main paper. The lead holds on every metric in every comparison. This includes Hausdorff distance, the one metric \ours{} does not lead in Table 2 of the main paper.

\begin{table}[h]
\centering
\caption{Autoregressive baselines inside and outside their published face caps, with \ours{} on the same meshes inside each cap and on the full sets. $n$ is the number of meshes scored per set.}
\label{tab:arcap}
\small
\setlength{\tabcolsep}{3.5pt}
\begin{tabular}{lcccccc@{\hspace{8pt}}cccc}
\toprule
& \multicolumn{2}{c}{$n$} & \multicolumn{4}{c}{Objaverse} & \multicolumn{4}{c}{Toys4K} \\
\cmidrule(lr){2-3}\cmidrule(lr){4-7}\cmidrule(lr){8-11}
Method & Obj. & Toys & \cdl{}$\downarrow$ & \cdo{}$\downarrow$ & \hd{}$\downarrow$ & \nc{}$\uparrow$ & \cdl{}$\downarrow$ & \cdo{}$\downarrow$ & \hd{}$\downarrow$ & \nc{}$\uparrow$ \\
\midrule
\multicolumn{11}{l}{\emph{MeshAnything, cap 800}} \\
within the cap & 15 & 131 & .02838 & .03655 & .1661 & .7537 & .03525 & .04798 & .1765 & .7754 \\
\quad \ours{}, same meshes & 15 & 131 & \textbf{.00615} & \textbf{.00872} & \textbf{.0693} & \textbf{.8950} & \textbf{.00563} & \textbf{.00809} & \textbf{.0649} & \textbf{.9296} \\
full set & 185 & 500 & .03088 & .04105 & .1854 & .7083 & .03624 & .05020 & .1766 & .7539 \\
\midrule
\multicolumn{11}{l}{\emph{MeshAnything V2, cap 1,600}} \\
within the cap & 25 & 215 & .01907 & .02611 & .1513 & .7867 & .03549 & .04858 & .1683 & .7667 \\
\quad \ours{}, same meshes & 25 & 215 & \textbf{.00596} & \textbf{.00859} & \textbf{.0691} & \textbf{.8856} & \textbf{.00544} & \textbf{.00788} & \textbf{.0579} & \textbf{.9251} \\
full set & 185 & 500 & .02951 & .03981 & .1858 & .7306 & .03750 & .05143 & .1806 & .7524 \\
\midrule
\multicolumn{11}{l}{\emph{BPT, cap 8,000}} \\
within the cap & 90 & 380 & .01561 & .02155 & .0991 & .8406 & .00866 & .01256 & .0617 & .8962 \\
\quad \ours{}, same meshes & 90 & 380 & \textbf{.00660} & \textbf{.00937} & \textbf{.0708} & \textbf{.8728} & \textbf{.00561} & \textbf{.00814} & \textbf{.0576} & \textbf{.9094} \\
full set & 185 & 500 & .02086 & .02858 & .1301 & .7963 & .01044 & .01482 & .0670 & .8898 \\
\midrule
\ours{}, full set & 185 & 500 & \textbf{.00649} & \textbf{.00933} & \textbf{.0789} & \textbf{.8529} & \textbf{.00560} & \textbf{.00814} & \textbf{.0666} & \textbf{.9069} \\
\bottomrule
\end{tabular}
\end{table}

TreeMeshGPT, DeepMesh and FastMesh publish no face cap, so there is no subset to restrict them to and they are scored on the full sets in Table 2 of the main paper.

\section{Additional Qualitative Results}
\label{app:qual_more}
\begin{figure}[p]
\centering
\includegraphics[width=\linewidth]{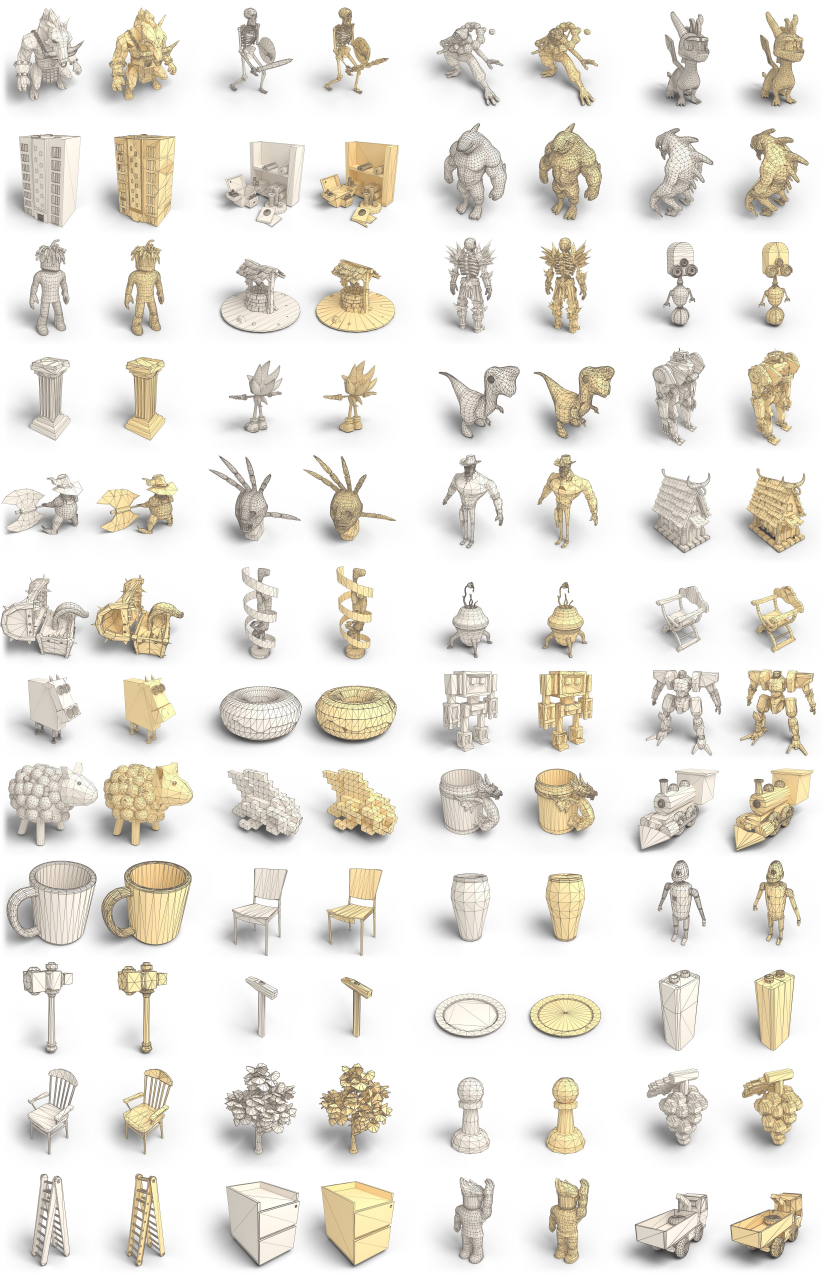}
\caption{Additional results of \ours{} on Objaverse and Toys4K. In each pair the reference artisan mesh is on the left in ivory and the generated mesh on the right in amber.}
\label{fig:qual_more}
\end{figure}

Figure~\ref{fig:qual_more} shows 48 further objects from Objaverse and Toys4K, each as the artist reference beside the mesh \ours{} generates from its point cloud, rendered as generated with their wireframes.

\begin{figure}[!t]
\centering
\includegraphics[width=\linewidth]{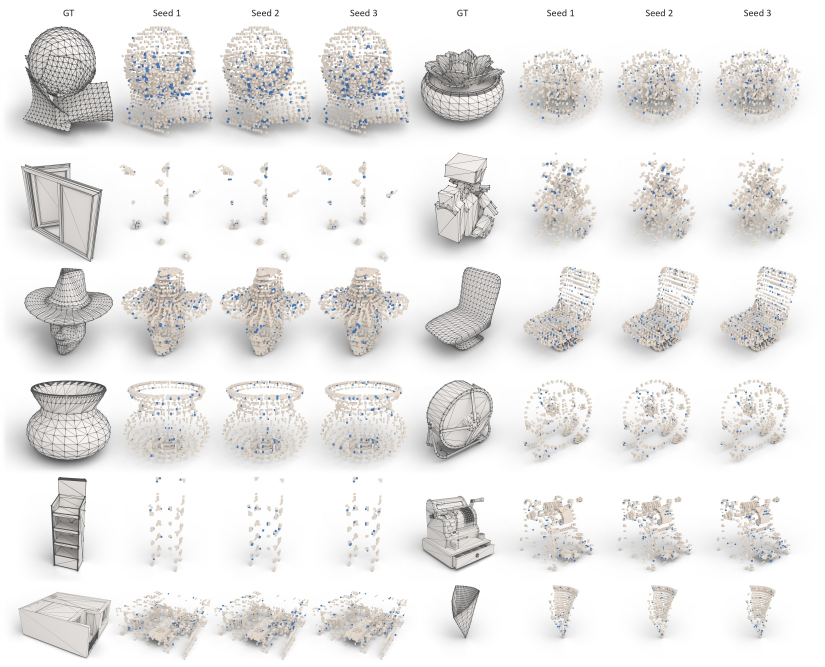}
\caption{Draws of the anchor generator. Each object shows the reference mesh (GT) and three draws of its $64^3$ latent voxel anchor at the reference vertex budget. Blue voxels differ from the other two draws.}
\label{fig:s1_seeds}
\end{figure}

\section{Limitations and Future Work}
\label{app:limit}
The VAEs bound the pipeline: their reconstruction is close but not exact, and everything downstream inherits that ceiling. Figure~\ref{fig:qual_more} shows two further limitations. Above 10K vertices, $512^3$ quantization can reduce mesh quality. The edge flow is teacher-forced on reference vertices but receives generated ones at inference, creating a domain gap, so it does better given the reference (Table~\ref{tab:main}, top block). Future work can improve vertex distribution matching and adapt the edge flow to generated vertices by training it on the vertex stage's output.

\section{Inference Cost}
\label{app:cost}
We time \ours{} stage by stage in its deployed setting, with the 50 vertex-flow steps and 12 edge-flow steps of Table 2 of the main paper, on one NVIDIA RTX PRO 6000 Blackwell Max-Q. The set is 24 Objaverse meshes with 500 to 4,600 reference vertex voxels at $512^3$, six per size bin, and \ours{} receives the reference vertex count. Model loading, compilation and autotuning are excluded: every stage processes the set twice in one process and only the second pass is timed. Input preparation is not counted. Table~\ref{tab:cost} gives the mean time per stage.

\ours{} generates a mesh in about 11\,s end to end, although its vertex and edge flows have 1.27B and 2.37B parameters, several times more than the flows of previous methods such as LATO.2 (162M and 221M, counted from its released checkpoints). Autoregressive methods, in comparison, often take minutes per mesh~\citep{fastmesh,nexus}. The vertex flow takes most of the time, and the total varies little with the mesh, since the compressed latent holds at most a few thousand tokens: it is 11.1\,s on the meshes with 500 to 1,000 vertex voxels and 12.3\,s on those with 3,400 to 4,600 (Table~\ref{tab:cost}). Scoring all vertex pairs in the connection head is quadratic in the vertex count, but it is not a concern in practice. With chunked scoring on the GPU, the connection head and the face enumeration take about 0.1\,s together on average.

\begin{table}[h]
\centering
\small
\caption{Generation time of \ours{} per mesh in seconds, means over 24 Objaverse meshes on one RTX PRO 6000 Blackwell Max-Q, in total and per bin of reference vertex voxels at $512^3$.}
\label{tab:cost}
\begin{tabular}{lc}
\toprule
Stage & Time (s) \\
\midrule
Anchor generator & 0.10 \\
Vertex stage: condition, 50-step vertex flow, VertexVAE decode & 9.45 \\
Edge stage: EdgeVAE encode, condition, 12-step edge flow, EdgeVAE decode & 1.79 \\
Connection head over all vertex pairs & 0.09 \\
Face enumeration & $<$0.01 \\
\midrule
Total & 11.4 \\
\quad 500--1,000 vertex voxels & 11.1 \\
\quad 1,000--2,000 vertex voxels & 11.1 \\
\quad 2,000--3,400 vertex voxels & 11.2 \\
\quad 3,400--4,600 vertex voxels & 12.3 \\
\midrule
Input preparation, not counted & 0.05 \\
\bottomrule
\end{tabular}
\end{table}

\section{Sampling the Anchor Generator}
\label{app:s1_vae}
The anchor generator is a VAE, so the anchor can be sampled rather than predicted. Its decoder takes either the posterior mean, the setting of Table 2 of the main paper, or a draw from the posterior, and decodes it in one forward pass. A user can regenerate the anchor with a different seed to suit their preference, as with the two flow stages. The VAE is a generative model by construction and has served generative tasks before~\citep{vae,palumbo2023mmvaeplus}. It fits the anchor well because the coarse geometry leaves little ambiguity, so one sampled pass replaces the many steps a flow would take. Table~\ref{tab:s1_vae} compares the two ways to decode it end to end: the posterior mean and a posterior sample.

\begin{table}[h]
\centering
\small
\caption{End-to-end results on 185 Objaverse meshes.}
\label{tab:s1_vae}
\begin{tabular}{lcccc}
\toprule
Anchor & \cdl{}$\downarrow$ & \cdo{}$\downarrow$ & \hd{}$\downarrow$ & \nc{}$\uparrow$ \\
\midrule
Posterior mean & .00649 & .00933 & .0789 & .8529 \\
Posterior sample & .00674 & .00964 & .0803 & .8516 \\
\bottomrule
\end{tabular}
\end{table}

The posterior mean is the expected anchor for the input, so it is the more stable choice. A posterior sample adds noise to the latent, and it is natural that it scores slightly worse: 2\% in Hausdorff distance and 3 to 4\% in Chamfer distance, with nearly the same normal consistency. The draws differ in a small share of their voxels, so sampling adds variation to the anchor without departing from it, and each draw places somewhat more voxels than the reference count because the decoder is trained toward completeness. Figure~\ref{fig:s1_seeds} shows three draws for 12 objects chosen for visible variation.

\end{document}